\def\year{2021}\relax
\documentclass[letterpaper]{article} 
\usepackage{aaai21}  
\usepackage{times}  
\usepackage{helvet} 
\usepackage{courier}  
\usepackage[hyphens]{url}  
\usepackage{graphicx} 
\usepackage{natbib}  
\usepackage{caption} 
\usepackage{booktabs}
\usepackage{multirow}
\usepackage{dsfont}
\usepackage[version=3]{mhchem}

\usepackage[T1]{fontenc}
\usepackage{amsmath,amssymb,amsfonts,mathrsfs,bm}
\usepackage{mathtools}
\usepackage{amsthm}
\usepackage[shortlabels]{enumitem}
\usepackage{graphicx}
\usepackage{epstopdf}
\usepackage{url}
\usepackage{colortbl}
\usepackage{multirow}
\usepackage{xcolor}
\usepackage{array}
\newcolumntype{L}[1]{>{\raggedright\let\newline\\\arraybackslash\hspace{0pt}}m{#1}}
\newcolumntype{C}[1]{>{\centering\let\newline\\\arraybackslash\hspace{0pt}}m{#1}}
\newcolumntype{R}[1]{>{\raggedleft\let\newline\\\arraybackslash\hspace{0pt}}m{#1}}

\makeatletter
\let\MYcaption\@makecaption
\makeatother
\usepackage[font=footnotesize]{subcaption}
\makeatletter
\let\@makecaption\MYcaption
\makeatother

\usepackage{xparse}



\usepackage{glossaries}
\newacronym{wrt}{w.r.t.}{with respect to}
\newacronym{RHS}{R.H.S.}{right-hand side}
\newacronym{LHS}{L.H.S.}{left-hand side}
\newacronym{iid}{i.i.d.}{independent and identically distributed}

\usepackage[capitalize]{cleveref}
\crefname{equation}{}{}
\Crefname{equation}{}{}
\crefname{claim}{claim}{claims}
\crefname{step}{step}{steps}
\crefname{line}{line}{lines}
\crefname{condition}{condition}{conditions}
\crefname{dmath}{}{}
\crefname{dseries}{}{}
\crefname{dgroup}{}{}
\crefname{Theorem}{Theorem}{Theorems}
\crefname{Corollary}{Corollary}{Corollaries}
\crefname{Proposition}{Proposition}{Propositions}
\crefname{Lemma}{Lemma}{Lemmas}
\crefname{Definition}{Definition}{Definitions}
\crefname{Example}{Example}{Examples}
\crefname{Assumption}{Assumption}{Assumptions}
\crefname{Remark}{Remark}{Remarks}
\crefname{Rem}{Remark}{Remarks}
\crefname{remarks}{Remarks}{Remarks}
\crefname{Exercise}{Exercise}{Exercises}
\crefname{Theorem_A}{Theorem}{Theorems}
\crefname{Corollary_A}{Corollary}{Corollaries}
\crefname{Proposition_A}{Proposition}{Propositions}
\crefname{Lemma_A}{Lemma}{Lemmas}
\crefname{Definition_A}{Definition}{Definitions}

\usepackage{algorithm,algorithmic}

\ifx\loadbreqn\undefined
	\relax
\else
	\usepackage{breqn} 
\fi

\ifx\renewtheorem\undefined
\ifx\useTheoremCounter\undefined
\newtheorem{Theorem}{Theorem}
\newtheorem{Corollary}{Corollary}
\newtheorem{Proposition}{Proposition}
\newtheorem{Lemma}{Lemma}
\else

\fi

\newtheorem{Assumption}{Assumption}

\fi

\theoremstyle{remark}

\theoremstyle{plain}

\newcommand{\Real}{\mathbb{R}}

\newcommand{\calL}{\mathcal{L}}

\newcommand{\calU}{\mathcal{U}}

\newcommand{\bc}{\mathbf{c}}

\newcommand{\boldf}{\mathbf{f}}

\newcommand{\bJ}{\mathbf{J}}

\newcommand{\boldm}{\mathbf{m}}
\newcommand{\bM}{\mathbf{M}}
\newcommand{\bn}{\mathbf{n}}

\newcommand{\bp}{\mathbf{p}}

\newcommand{\br}{\mathbf{r}}

\newcommand{\bs}{\mathbf{s}}
\newcommand{\bS}{\mathbf{S}}

\newcommand{\by}{\mathbf{y}}

\newcommand{\bz}{\mathbf{z}}

\DeclareSymbolFont{bsfletters}{OT1}{cmss}{bx}{n}
\DeclareSymbolFont{ssfletters}{OT1}{cmss}{m}{n}
\DeclareMathSymbol{\bsfGamma}{0}{bsfletters}{'000}
\DeclareMathSymbol{\ssfGamma}{0}{ssfletters}{'000}
\DeclareMathSymbol{\bsfDelta}{0}{bsfletters}{'001}
\DeclareMathSymbol{\ssfDelta}{0}{ssfletters}{'001}
\DeclareMathSymbol{\bsfTheta}{0}{bsfletters}{'002}
\DeclareMathSymbol{\ssfTheta}{0}{ssfletters}{'002}
\DeclareMathSymbol{\bsfLambda}{0}{bsfletters}{'003}
\DeclareMathSymbol{\ssfLambda}{0}{ssfletters}{'003}
\DeclareMathSymbol{\bsfXi}{0}{bsfletters}{'004}
\DeclareMathSymbol{\ssfXi}{0}{ssfletters}{'004}
\DeclareMathSymbol{\bsfPi}{0}{bsfletters}{'005}
\DeclareMathSymbol{\ssfPi}{0}{ssfletters}{'005}
\DeclareMathSymbol{\bsfSigma}{0}{bsfletters}{'006}
\DeclareMathSymbol{\ssfSigma}{0}{ssfletters}{'006}
\DeclareMathSymbol{\bsfUpsilon}{0}{bsfletters}{'007}
\DeclareMathSymbol{\ssfUpsilon}{0}{ssfletters}{'007}
\DeclareMathSymbol{\bsfPhi}{0}{bsfletters}{'010}
\DeclareMathSymbol{\ssfPhi}{0}{ssfletters}{'010}
\DeclareMathSymbol{\bsfPsi}{0}{bsfletters}{'011}
\DeclareMathSymbol{\ssfPsi}{0}{ssfletters}{'011}
\DeclareMathSymbol{\bsfOmega}{0}{bsfletters}{'012}
\DeclareMathSymbol{\ssfOmega}{0}{ssfletters}{'012}

\newcommand{\bzeta}{\bm{\zeta}}

\newcommand{\bphi}{\bm{\phi}}

\DeclareMathOperator*{\argmax}{arg\,max}

\DeclareMathOperator{\rank}{rank}

\DeclareMathOperator{\sign}{sign}

\DeclarePairedDelimiter\braces{\{}{\}}

\newcommand{\qednew}{\nobreak \ifvmode \relax \else
      \ifdim\lastskip<1.5em \hskip-\lastskip
      \hskip1.5em plus0em minus0.5em \fi \nobreak
      \vrule height0.75em width0.5em depth0.25em\fi}

\newcommand{\nn}{\nonumber\\}

\newcommand{\T}{^{\intercal}}

\newcommand{\ceil}[1]{\left\lceil{#1}\right\rceil}
\newcommand{\floor}[1]{\left\lfloor{#1}\right\rfloor}

\newcommand{\cond}[2]{\left. {#1}\, \middle| \, {#2} \right.}

\DeclareDocumentCommand \P { g d() g } {%
	\IfNoValueTF {#3} 
	{%
		\IfNoValueTF {#1} 
		{%
			\IfNoValueTF {#2}
			{%
				\mathbb{P}%
			}%
			{%
				\mathbb{P}\left({#2}\right)%
			}%
		}%
		{%
			\IfNoValueTF {#2}
			{%
				\mathbb{P}_{#1}%
			}%
			{%
				\mathbb{P}_{#1}\left({#2}\right)%
			}%
		}%
	}%
	{%
		\IfNoValueTF {#1} 
		{%
			\mathbb{P}\left(\cond{#2}{#3}\right)%
		}%
		{%
			\mathbb{P}_{#1}\left(\cond{#2}{#3}\right)%
		}%
	}%
}

\DeclareDocumentCommand \E { g o g } {%
	\IfNoValueTF {#3} 
	{%
		\IfNoValueTF {#1} 
		{%
			\IfNoValueTF {#2}
			{%
				\mathbb{E}%
			}%
			{%
				\mathbb{E}\left[{#2}\right]%
			}%
		}%
		{%
			\IfNoValueTF {#2}
			{%
				\mathbb{E}_{#1}%
			}%
			{%
				\mathbb{E}_{#1}\left[{#2}\right]%
			}%
		}%
	}%
	{%
		\IfNoValueTF {#1} 
		{%
			\mathbb{E}\left[\cond{#2}{#3}\right]%
		}%
		{%
			\mathbb{E}_{#1}\left[\cond{#2}{#3}\right]%
		}%
	}%
}

\definecolor{gray90}{gray}{0.9}

\ifx\nohighlights\undefined
	\newcommand{\red}[1]{{\color{red} #1}}
	
	\newcommand{\msout}[1]{\text{\color{green} \sout{\ensuremath{#1}}}}
	\newcommand{\del}[1]{{\color{green}\ifmmode \msout{#1}\else\sout{#1}\fi}}
\else
	\newcommand{\red}[1]{#1}
	
	\newcommand{\msout}[1]{#1}
	\newcommand{\del}[1]{#1}
\fi

\graphicspath{{./Figures/}} 
\ifx\diagnoselabel\undefined
	\relax
\else
	\makeatletter
	 \def\@testdef #1#2#3{%
		 \def\reserved@a{#3}\expandafter \ifx \csname #1@#2\endcsname
		\reserved@a  \else
	 \typeout{^^Jlabel #2 changed:^^J%
	 \meaning\reserved@a^^J%
	 \expandafter\meaning\csname #1@#2\endcsname^^J}%
	 \@tempswatrue \fi}
	\makeatother
\fi

\newcommand{\VI}{\mathrm{VI}}

\title{Error-Correcting Output Codes with Ensemble Diversity for Robust Learning in Neural Networks}
\author{
    Yang~Song$^*$, Qiyu~Kang\thanks{Two authors contributed equally to this work.}, and Wee~Peng~Tay \\
}
\affiliations{
    Nanyang Technological University, 
    50 Nanyang Ave, Singapore 639798 \\
    songy@ntu.edu.sg, kang0080@e.ntu.edu.sg, wptay@ntu.edu.sg 

}
\begin{document}
\maketitle

\begin{abstract}

Though deep learning has been applied successfully in many scenarios, malicious inputs with human-imperceptible perturbations can make it vulnerable in real applications. This paper proposes an error-correcting neural network (ECNN) that combines a set of binary classifiers to  combat adversarial examples in the multi-class classification problem. To build an ECNN, we propose to design a code matrix  so that the minimum Hamming distance between any two rows (i.e., two codewords) and the minimum shared information distance between any two columns (i.e., two partitions of class labels) are simultaneously maximized. Maximizing row distances can increase the system fault tolerance while maximizing column distances helps increase the diversity between binary classifiers. We propose an end-to-end training method for our ECNN, which allows further improvement of the diversity between binary classifiers. The end-to-end training renders our proposed ECNN different from the traditional error-correcting output code (ECOC) based methods that train binary classifiers independently. 
ECNN is complementary to other existing defense approaches such as adversarial training and can be applied in conjunction with them. 
We empirically demonstrate that our proposed ECNN is effective against the state-of-the-art white-box and black-box attacks on several datasets while maintaining good classification accuracy on normal examples. 
\end{abstract}

\section*{Introduction}
\label{sect:intro}
Deep learning has been widely and successfully applied in many tasks such as image classification \cite{KrizhevskyNIPS2012, LecunPIEEE1998}, speech recognition \cite{HintonSPM2012}, and natural language processing \cite{AndorACL2016}. 
However, recent works \cite{SzegedyICLR2013} showed that the original images can be modified by an adversary with human-imperceptible perturbations so that the deep neural networks (DNNs) are fooled into mis-classifying them. 
To mitigate the effect of adversarial attacks, many defense approaches have been proposed. Generally speaking, they fall into three categories: 
\begin{enumerate}[leftmargin=\parindent,align=left,labelwidth=\parindent,labelsep=0pt]
\item adversarial training, which augments the training data with adversarial examples \cite{SzegedyICLR2013, GoodICLR2015},
\item \label{it:2}modifying the DNN or training procedure, e.g., defensive distillation \cite{PapISSP2016}, and  
\item post-training defenses, which attempt to remove the adversarial noise from the input examples \cite{HendrycksICLR2017, MengCCS2017,SamICLR2018} in the testing phase. 
\end{enumerate}
In this paper, we propose an approach for the second category. Most defense approaches of this type focus on robustifying a single network, while a few works have adopted ensemble methods \cite{Abb2017, Xu2018NDSS,PangICML2019,Sen2020empir}. These ensemble methods build a new classifier consisting of several base classifiers that are assigned with the same classification task. Promoting the diversity among the base classifiers during training is essential to prevent adversarial examples from transferring between them, since the adversarial examples crafted for one classifier may also fool the others. 

The use of error correcting output codes (ECOC) \cite{DietJAIR1994,CarcTEC2008} differs from the above mentioned ensemble methods by assigning each class an unique codeword. This forms a pre-defined code matrix. For an illustration, see \cref{tab:ecn4}, which shows an example of a code matrix for three classes with each class being represented by four bits. Each column splits the original classes into two meta-classes, meta-class `0' and meta-class `1'. A binary classifier is then learned independently for each column of the code matrix. To classify a new sample, all binary classifiers are evaluated to obtain a binary string. Finally, the method assigns the class whose codeword is closest to the obtained binary string to the sample. 
In the literature, ECOC is mostly used with decision trees or shallow neural networks. 

\begin{table}[t]
\centering
\small
\begin{tabular}{ccccc}
\toprule
Class & Net 0 & Net 1 & Net 2 & Net 3 \\ 
\midrule
0 & 1     & 0     & 1     & 0 \\ 
1 & 1     & 1     & 0     & 1 \\ 
2 & 0     & 0     & 0     & 1 \\ 
\bottomrule
\end{tabular}
\caption{Example of a $3\times 4$ code matrix.}\label{tab:ecn4}
\end{table}

In this paper, we utilize the concept of ECOC in a deep neural network, which we call error-correcting neural network (ECNN). In traditional ECOC, a code matrix is generated in such a way that the minimum Hamming distance between any two rows is maximized. Maximizing the row distance creates sufficient redundant error-correcting bits, thus enhancing the classifier's error-tolerant ability. However, it is possible for an adversary to design adversarial examples that trick most of the binary classifiers since they are not fully independent of each other. Therefore, to mitigate this effect, the code matrix should be designed so that the binary tasks are as different from each other as possible. When designing a code matrix for ECNN, we attempt to separate columns (binary tasks) by maximizing the minimum shared information distance \cite{meilua2003comparing} between any two columns. Maximizing the column distance inherently promotes the diversity between binary classifiers. This is essential to prevent the adversarial examples crafted for one binary classifier transferring to the other binary classifiers. 

After our preliminary work, we became aware of a recent independent work \cite{VermaNIPS2019}, which designs a DNN using error-correcting codes and shares similar concepts with our proposed approach.  The main difference to our work is that our encoder, i.e., all the binary classifiers are trained jointly whereas \cite{VermaNIPS2019} splits the binary classifiers into several groups, with each being trained separately. 
The work \cite{VermaNIPS2019} only forces a pre-training diversity using its code matrix whereas ECNN further includes a diversity promoting regularizer during training. This novelty improves the testing accuracy of ECNN compared to \cite{VermaNIPS2019}. 

During training of ECNN, all binary classifiers are jointly trained and their outputs, i.e., the predicted meta-class classification probabilities, are concatenated and fed into a decoder to obtain the predicted classification probabilities. We propose an end-to-end training method that allows for exploiting the interaction between binary classifiers, thus further improving the diversity between them. Our main contributions are summarized as follows:
\begin{enumerate}
\item We apply error-correcting codes to build a DNN for classification and propose an end-to-end training method. For illustration, we focus our discussion on the problem of robust image classification.
\item We provide theoretical analysis that helps to guide the design of ECNN, including the choice of activation functions.
\item In our experiments,  we test ECNN on several widely used datasets MNIST \cite{LeCunMNIST}, CIFAR-10 and CIFAR-100 \cite{KriTR2009} under several well-known adversarial attacks. We demonstrate empirically that ECNN is robust against adversarial white-box attacks with improvement in classification accuracy of adversarial examples of up to $14.8$ and $17.4$ percentage points compared to another current state-of-the-art ensemble method \cite{VermaNIPS2019} on MNIST and CIFAR-10, respectively, while ECNN uses $22.2\%$ more parameters than \cite{VermaNIPS2019} on MNIST and $78.8\%$ {\em less} parameters than \cite{VermaNIPS2019} on CIFAR-10. 
\item We also test ECNN on the German Traffic Sign Recognition Benchmark (GTSRB) \cite{Stallkamp2012} under black-box setting and show that ECNN outperforms the baseline on both normal and adversarial examples.
\item When combined with adversarial training, ECNN further improves its robustness, with improvement in correct classification of adversarial examples by about $18$ percentage points compared to pure adversarial training.
\end{enumerate}
Furthermore, we show how to generalize the binary classifiers used in ECNN to $q$-ary classifiers using a $q$-ary code matrix.

The rest of this paper is organized as follows. 
Firstly, we present our ECNN framework, its training strategy and some theoretical analysis of its properties. Then, we present extensive experimental results, and we conclude in the last section. We refer interested readers to the supplementary material for a more detailed account of several recent works that defend against adversarial examples using ensembles of models \cite{Abb2017, Xu2018NDSS, PangICML2019,VermaNIPS2019,Sen2020empir} and some popular adversarial attacks \cite{GoodICLR2015,KurakinICLR2017a,MadryICLR2018,PapernotESSP,CarliniISSP2017} that are used to verify the robustness of our proposed ECNN. The proofs for all lemmas in this paper are given in the supplementary material. 
For easier reference, we summarize some of the commonly-used symbols in \cref{tab:notations}.
\begin{table}[t]
\centering
\small
\begin{tabular}{cl}
\toprule
Notation & Definition \\
\midrule
$N$ & number of binary classifiers\\
$g_{\theta_n}(\cdot)$ & feature extraction function at the $n$-th classifier \\
${\bf f}_n$ & feature vector ${\bf f}_n=g_{\theta_n}(x)$ \\
$F$ & dimension of feature vector ${\bf f}_n$\\
$h_n(\cdot)$ & prediction function at the $n$-th binary classifier  \\
$z_n$ & encoder's outputs $z_n=h_n({\bf f}_n)$  \\
$\bphi_n$ & linear form of $h_n$, i.e., $z_n=\bphi_n {\bf f}_n$ \\
$K$ & number of input samples $x\in\{x_0,\ldots,x_{K-1}\}$\\
${\bf f}_{n,x}$ & ${\bf f}_n$ depending on input sample $x$\\
${\bf f}_n^{y(x)}$ & principal feature vector associated with class $y(x)$ \\
$\bn_{n,x}$ & random perturbation  ${\bf f}_{n,x}={\bf f}_n^{y(x)}+\bn_{n,x}$ \\
$\nu(\cdot)$ & logistic function \\
$\sigma(\cdot)$ & softmax function \\
$y_n(x)$ & meta-class $y_n(x)=\bM(y(x),n)$\\
\bottomrule
\end{tabular}
\caption{Summary of commonly-used symbols.}\label{tab:notations}
\end{table}

\section*{Error-Correcting Neural Network}
\label{sect:err_cor}

In this section, we present the architectures of the encoder and the decoder in ECNN, the training strategy and the way to generate the code matrix. We provide theoretical results that help to guide us in the design of ECNN. Finally, we show how to extend to $q$-ary classifiers in the encoder, where $q>2$.

\begin{figure}[t]
\centering
\includegraphics[width=0.42\textwidth, height=1.1in]{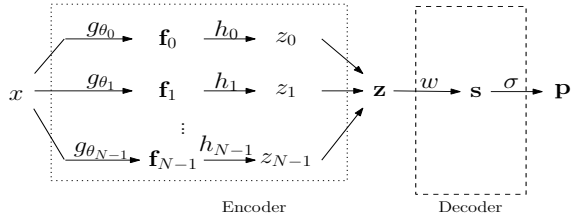}
\caption{ECNN architecture.}
\label{fig:ecn}
\end{figure}

The overall architecture of ECNN is shown in \cref{fig:ecn}. Consider a $M$-ary classification problem. A $M\times N$ binary code matrix $\bM$, where $N\geq 1$, encodes each class with a $N$-bit codeword. For each $n=0,\ldots,N-1$, the $n$-th bits of all the $M$ codewords define a binary meta-classification problem, with $y_n(x)=\bM(y(x),n)\in\{0,1\}$ being the meta-class of sample $x$ if $y(x)$ is the label of $x$. 
For example, in the code matrix shown in \cref{tab:ecn4}, the first bits or ``Net 0'' as indicated in the table correspond to a binary classification problem that distinguishes classes 0 or 1 from class 2. We learn a binary classifier corresponding to each binary meta-classification problem and combine their outputs together. The collection of these binary classifiers is called the \emph{encoder} in our architecture. We call each binary classifier in the encoder a \emph{meta classifier}. The encoder is then followed by a \emph{decoder} whose function is to infer which of the $M$ classes the sample belongs to. In the following, we describe both the encoder and decoder in detail. Let $y(x)$ be the label of sample $x$. We suppose that a training set $\braces{(x_k,y(x_k)) : k = 0,\ldots, K-1}$ is available. In our discussions, we append a subscript $x_k$ to a quantity (e.g., $\boldf_{n,x_k}$ in place of $\boldf_n$) if we wish to emphasize its dependence on the sample $x_k$.

\subsection*{Encoder} 

The encoder contains $N$ composite functions $\sigma\circ h_n \circ g_{\theta_n}$, for $n=0,\ldots,N-1$, each corresponding to a binary classifier. The function $g_{\theta_n}$ extract feature vectors ${\boldf}_n = g_{\theta_n}(x)\in\Real^F$ from the input $x$, $h_n$ is a prediction function that outputs $z_n=h_n({\boldf}_n)\in\Real^1$, which can be interpreted as probabilities after normalization. 

We choose $h_n$ to be a simple linear function, i.e., $z_n=\bphi_n{\boldf}_n$, for $n=0,\ldots,N-1$. Furthermore, in our final architecture, we set $\bphi_n=\bphi$ for all $n=0,\ldots,N-1$. The reason for our choice is explained later in the next subsection where we introduce the concept of ensemble diversity. 
Let $\bM\in\{0,1\}^{M\times N}$ be the given code matrix
. Denote the loss function for the encoder as $\ell(z_n,y_n(x))$. The encoder can be formulated as an optimization problem:
\begin{align}\label{eq:enc}
\min_{\{\theta_n,\bphi_n\}_{n=0}^{N-1}}&\ \frac{1}{N(N-1)K} \sum_{k=0}^{K-1} \sum_{n=0}^{N-1} 
\ell(z_{n,x_k},y_n(x_k))  \\
{\rm s.t.}&~ \text{for $k=0,\ldots,K-1$ and $n=0,\ldots,N-1$}, \nn
&\ z_{n,x_k}= \bphi_n  g_{\theta_n}(x_k). \nonumber
\end{align}

 The loss function $\ell(z_n,y_n(x))$ would be
\begin{align}\label{eqn:loss}
&\max(0, 1-z_{n}(2y_n(x)-1)),\ \text{for hinge loss,} \\
&-{\bf 1}_{y_n(x)}\T\log\left(\bzeta_{n}\right) ,\ \text{for cross entropy loss},
\end{align}
where ${\bf 1}_{y_n(x)}$ is the one-hot encoding of the meta-class $y_n(x)$, i.e., a vector with 1 at the $y_n(x)$-th entry and 0 for all the other entries and $\bzeta_n=[1-\nu(z_n),\nu(z_n)]\T$ with $\nu(\cdot)$ being the logistic function. 
Finally, the logits $z_n$ are concatenated into $\bz=\left[z_0,\ldots,z_{N-1}\right]\T$, which is the input to the decoder in ECNN.

\subsection*{Ensemble Diversity}\label{subsec:ens_div} 
To mitigate against adversarial attacks, each feature vector ${\boldf}_n$, $n=0,\ldots,N-1$, should only be instrumental in its own binary classifier's prediction while being insensitive to the other binary classifiers' predictions. However, it is difficult to directly define a difference measurement between features. One possible way is to ensure linear independence between them and measure the independence using singular values. This operation is computationally costly, making it unsuitable for training in neural networks. Furthermore, even if feature vectors are linearly independent, it may very well happen that $\boldf_i$ is just a permutation of $\boldf_j$ for $i\ne j$. The use of $L_p$ distance to measure differences in feature vectors is therefore inappropriate. We choose to promote ensemble diversity in terms of the output probability of each binary classifier by solving the following optimization problem: 
\begin{align} \label{eq:div1}
\max_{\{\bphi_i\}_{i=0}^{N-1}} &~\frac{1}{N(N-1)K}\sum_{k=0}^{K-1}\sum_{i\neq j}
-\left[1-\nu\left(\bphi_i {\boldf}_{j,x_k}\right),\right. \nonumber\\
&~\left.\nu\left(\bphi_i {\boldf}_{j,x_k}\right)\right] \log\left(\left[1-\nu\left(\bphi_i {\boldf}_{j,x_k}\right),\nu\left(\bphi_i {\boldf}_{j,x_k}\right)\right]\T\right), \\
{\rm s.t.} &~ {\boldf}_{j,x_k}=g_{\theta_j}(x_k),\ j=0,\ldots,N-1,\nonumber \\
&~ k=0,\ldots,K-1. \nonumber
\end{align}

Assuming that the input $x$ is drawn from a distribution, let $\boldf^{y}_n$ be the expected feature vector of class $y$ generated by the $n$-th classifier in the encoder. We call this the principal feature vector of class $y$. Then for any input $x$, we have
\begin{align}\label{pri_feature}
\boldf_{n,x} = \boldf^{y(x)}_n + \bn_{n,x},
\end{align}
where $\bn_{n,x}$ is a zero-mean random perturbation. We make the following assumption.
\begin{Assumption}\label{assumpt:ac}
  The random vectors $\{\bn_{n,x_k} : k=0,\ldots,K-1\}$ have a joint distribution absolutely continuous \gls{wrt} Lebesgue measure.
\end{Assumption}

The above assumption is satisfied if the training samples $\{x_k : k=0,\ldots,K-1\}$ are drawn \gls{iid} from a distribution absolutely continuous \gls{wrt} Lebesgue measure.

%
%

\begin{Lemma}\label{lem:min}
Suppose that $\{\theta_n\}_{n=0}^{N-1}$ satisfy \cref{assumpt:ac}. Then the following statements hold with probability one:
\begin{enumerate}[(a)]
\item \label{it:lem2_a}  Suppose $\theta_n=\theta$ for all $n=0,\ldots,N-1$. There exists $\{\bphi_n : n=0,\ldots,N-1\}$ such that the loss of \cref{eq:enc} is  arbitrarily small if $K \leq F$.
\item \label{it:lem2_b} Suppose $\bphi_n=\bphi$ for all $n=0,\ldots,N-1$. There exists a $\bphi$ such that the loss of \cref{eq:enc} is arbitrarily small if $NK\leq F$.
\item \label{it:lem2_c}  Suppose $N>1$. Any feasible solution of \cref{eq:enc} cannot have $\theta_n=\theta$ and $\bphi_n=\bphi$ for some $\theta$ and $\bphi$, and for all $n=0,\ldots,N-1$.  
\end{enumerate}
\end{Lemma}

\begin{Lemma}\label{lem:bilinear}
  Suppose that $\{\theta_n\}_{n=0}^{N-1}$ satisfy \cref{assumpt:ac}. Suppose further that $\{\theta_n\}_{n=0}^{N-1}$ are chosen so that for each $n=0,\ldots,N-1$ and $y=0,\ldots,M-1$, $\boldf_n^y=\bS_n\boldf^y$ for some $\boldf^y\in\Real^F$, where $\bS_n$ is a permutation matrix. 
\begin{enumerate}[(a)]
\item There exist linear transformations $\{\bphi_n\}_{n=0}^{N-1}$ such that the loss of \cref{eq:enc} is arbitrarily small.
\item If $NM\ge F(M+1)$ and $\bphi_n$ are constrained to be the same for all $n=0,\ldots,N-1$, then for almost every (\gls{wrt} Lebesgue measure) $\{\boldf^y\}_{y=0}^{M-1}$, there is no feasible solution to \cref{eq:enc}. 
\end{enumerate}
\end{Lemma}

\begin{Lemma}\label{lem:highPr}
Suppose that $\{\theta_n\}_{n=0}^{N-1}$ satisfy \cref{assumpt:ac}, and $\braces{\boldf^y_n : y=0,\ldots,M-1, n=0,\ldots,N-1}$ are linearly independent. For all $n=0,\ldots,N-1$, $\bphi_n$ are constrained to be the same. Then, the $n$-th binary classifier in the encoder classifies the sample $x_k$, $k=0,\ldots,K-1$, correctly if $\|\bn_{n,x_k}\|_2$ is sufficiently small.
\end{Lemma}

%
%

\cref{lem:min} shows that if we constrain either the parameters $\theta_n$ or transforms $\bphi_n$ (but not both) in the encoder to be identical across binary classifiers, then it is still possible to achieve arbitrarily small loss. \cref{lem:bilinear} suggests that if we do not constrain the transforms $\bphi_n$ to be the same, then optimizing \cref{eq:enc} may produce a solution where the features $\boldf^{y(x)}_n$ for different binary classifiers $n=0,\ldots,N-1$ are permutations of each other for the same sample $x$. This is clearly undesirable as any adversarial attack on a particular binary classifier can then translate to the other classifiers. On the other hand, \cref{lem:highPr} suggests that if we constrain the transforms $\bphi_n$ to be the same, and choose the parameters $\theta_n$ to make $\braces{\boldf^y_n : y=0,\ldots,M-1, n=0,\ldots,N-1}$ linearly independent, then high classification accuracy is still achievable if the input sample does not deviate too much from the mean. Our results thus suggest that a good strategy is to share the transform $\bphi$ across all the prediction functions. Then, \cref{eq:div1} becomes 
\begin{align}\label{eq:div2}
\max_{\bphi}&~\frac{1}{NK}\sum_{k=0}^{K-1}\sum_{n=0}^{N-1} -\bzeta_{n,x_k}\T\log(\bzeta_{n,x_k}), \\
{\rm s.t.} &~ \text{for } n=0,\ldots,N-1,\ k=0,\ldots,K-1, \nonumber \\
&~ \bzeta_{n,x_k}=\left[1-\nu\left(\bphi {\boldf}_{n,x_k}\right),\nu\left(\bphi {\boldf}_{n,x_k}\right)\right]\T,\nonumber \\
&~ {\boldf}_{n,x_k}=g_{\theta_n}(x_k).\nonumber
\end{align}

\subsection*{Joint Optimization}
The joint optimization that combines \cref{eq:enc} and \cref{eq:div2} can be formulated as
\begin{align}\label{eq:enc_dec}
\min_{\{\theta_n\}_{n=0}^{N-1},\bphi}&~ 
\frac{1}{NK}\sum_{k=0}^{K-1} \sum_{n=0}^{N-1} 
\ell(z_{n,x_k},y_n(x_k)) \nonumber\\
&~~~ - \gamma\bzeta_{n,x_k}\log\left(\bzeta_{n,x_k}\right), \\
{\rm s.t.} &~ \text{for } n=0,\ldots,N-1,\ k=0,\ldots,K-1,\nn
&~ z_{n,x_k}=\bphi g_{\theta_n}(x_k), \nonumber
\end{align}
where $\gamma\geq 0$ is a configurable weight. The joint optimization \cref{eq:enc_dec} enables end-to-end training for ECNN.

Note that if we use cross entropy for $\ell(z_{n},y_n(x))$, we convert one-hot labels ${\bf 1}_{y_n(x_k)}$ to soft labels ${\bf 1}_{y_n(x_k)}-\gamma \bzeta_{n,x_k}$ for all $n,k$ by optimizing \cref{eq:enc_dec}. This is also known as label smoothing \cite{Sze2016}, which is beneficial in improving the adversarial robustness in ECNN. The following \cref{lem:solution} guides us in the choice of $\gamma$ given the smoothed probabilities $\{\bzeta_{n,x_k}(y_n(x_k)) : n=0,\ldots,N-1, k=0,\ldots,K-1\}$. 
\begin{Lemma}\label{lem:solution}
The optimal solution of \cref{eq:enc_dec}, where we use cross entropy for $\ell(z_{n},y_n(x))$,  satisfies $\frac{1}{\bzeta_{n,x_k}(y_n(x_k))}=\gamma \log\frac{\bzeta_{n,x_k}(y_n(x_k))}{1-\bzeta_{n,x_k}(y_n(x_k))}$ for all $n=0,\ldots,N-1, k=0,\ldots,K-1$.
\end{Lemma}

\subsection*{Decoder} 
\label{subsect:decoder}
The decoder involves a simple comparison with the code matrix $\bM$. To enable the use of back-propagation, our decoder uses continuous  relaxation: we compute the correlation between the real-valued string $\tanh(z)$ and each class's codeword (after scaling
$[-1,1]$), and the class having the maximum correlation is assigned as the output label. Mathematically, 
the decoder is a composite function $\sigma \circ w$. The function $w$ first scales the logits $\bz$ to $[-1,1]$ using $\tanh$ function and then computes the inner products between the scaled logits and each row in $2\bM-\mathds{1}$, i.e.,
$\bs=w(\tanh(\bz))=(2\bM-\mathds{1}) \tanh(\bz) \in\Real^M$, where $\mathds{1}$ denotes a matrix with all its entries being 1. The  $\sigma(\cdot)$ is the softmax function which maps $\bs$ to the prediction probabilities $\bp={\sigma}\left((2\bM-\mathds{1}) \tanh(\bz)\right)\in\Real^M$.

\subsection*{Code Matrix Design}\label{ss:cmdesign} 
Let $\br_i$ be the $i$-th codeword of code matrix $\bM$, and  $H(\br_i, \br_j)$ be the Hamming distance between the $i$-th and $j$-th codewords of the code matrix, i.e., the number of bit positions where they differ. The minimum Hamming distance of code matrix $\bM$ is then given by $d_{H}(\bM) = \min_{i\ne j} H(\br_i, \br_j)$. A code matrix with larger $d_{H}(\bM)$ is preferred since it can correct more errors, resulting in better classification performance. However, if we only consider maximizing $d_{H}(\bM)$ when designing $\bM$, the two columns of the matrix may lead to the corresponding binary classifiers performing the same classification task even though they have different bits. For a concrete illustration, consider again the code matrix example in \cref{tab:ecn4}. The last two columns in \cref{tab:ecn4} are different, while the corresponding ``Net 2'' and ``Net 3'' are essentially performing the same task: classifying the classes to set $\{0\}$ or $\{1,2\}$. An adversarial example generated by an untargeted attack that fools ``Net 2'' will also fool ``Net 3''. To further promote the diversity between binary classifiers, we therefore include column diversity when designing the code matrix. Specifically, we view each column of the code matrix as partitioning the $M$ classes into clusters, and measure the difference between two columns using the variation of information (VI) metric \cite{meilua2003comparing}. VI, which is a criterion for comparing the difference between two binary partitions, measures the amount of information lost and gained in changing from one clustering to another clustering \cite{meilua2003comparing}. Each column of the ECNN code matrix can be interpreted to be a binary cluster. We denote the $n$-th column of code matrix $\bM$ as $\bc_n$, a set of classes that belong to meta-class $k$ in the $n$-th column as $C_n^k$, where $n =0,\ldots,N-1$ and  $k = 0,1$.
The VI distance defined in \cite{meilua2003comparing} between $\bc_m$ and $\bc_n$, i.e., $\VI(\bc_m,\bc_n)$, is
\begin{align}\label{eq:vi}
& - \sum_{k = 0}^{1} s_m(k) \log s_m(k) - \sum_{k' = 0}^{1} s_n(k') \log s_n(k') \nn
&\hspace{-.2in}- 2\sum_{k=0}^{1}\sum_{k'=0}^{1} s_{m,n}(k,k')\log \frac{s_{m,n}(k,k')}{s_m(k)s_n(k')},
\end{align}
where $s_{m,n}(k,k') = \frac{\left|C_m^k \cap C_n^{k'}\right|}{M}$,  $s_m(k) = \frac{\left|C_m^k\right|}{M}$, $s_n(k') = \frac{\left|C_n^{k'} \right|}{M}$, and $|A|$ denotes the cardinality of the set $A$.  The minimum VI distance of code matrix $\bM$ is then given by $d_{\VI}(\bM) = \min_{m\ne n} \VI(\bc_m, \bc_n)$.  

The code matrix is designed to $\max_{\bM} d_{H}(\bM) + d_{\VI}(\bM)$,
which is however a NP-complete problem \cite{pujol2006discriminant}. We adopt simulated annealing \cite{kirkpatrick1983optimization,gamal1987using} to solve this optimization problem heuristically, where the energy function is set as:
\begin{align}\label{eq:Mobj_alternative}
\min_{\bM} \sum_{i\ne j} H(\br_i, \br_j)^{-2} + \eta\sum_{m\ne n}\VI(\bc_m,\bc_n)^{-2}, 	
\end{align}
and $\eta$ is chosen such that the two summations are roughly equally weighted. This is a common technique used in simulated annealing \cite{gamal1987using} as bit changes not involving the minimum distance pairs are not reflected in the energy function if it is set as $\max_{\bM} d_{H}(\bM) + d_{\VI}(\bM)$.

\subsection*{$q$-ary ECNN} 
A natural extension is to use a general $q$-ary code matrix with each meta classifier performing a $q$-ary classification problem. 
Since $\max{d_{H}(\bM)}$ for a $q$-ary $\bM$ is no less than $\max{d_{H}(\bM)}$ for a binary $\bM$, using a $q$-ary $\bM$ with $q>2$ has better error-correcting capacity than using a binary $\bM$. Hence, better classification accuracy on normal images is expected as $q$ increases. On the other hand, we show in \cref{lem:I} below that more information about the original classes is revealed in each $q$-ary classifier as $q$ increases, thus rendering less adversarial robustness.

\begin{Lemma}\label{lem:I}
Suppose $\by=[0,\ldots,M-1]\T$ and $\bc$ is a column of a $q$-ary code matrix $\bM$. The mutual information between $\by$ and $\bc$ can be defined, analogously to \cref{eq:vi}, as \begin{align}\label{eq:I}
I(\by,\bc)=\sum_{k=0}^{q-1}\sum_{\ell=0}^{M-1} s(k,\ell) \log \frac{s(k,\ell)}{s(k)s(\ell)},
\end{align}
where $s(k,\ell) = {\left| C^{k}\cap \left\{ \ell \right\}\right|}/{M}$, $s(k) = {\left|C^k\right|}/{M}$ with $C^k$ being the set of meta-classes in $\bc$, and $s(\ell) = {\left|\left\{ \ell \right\}\right|}/{M}=1/{M}$ that belong to meta-class $k$. Then, $\max_\bc I(\by,\bc)$ is an increasing function of $q$.
\end{Lemma}

The training process of a $q$-ary ECNN is the same as that of a binary ECNN introduced in the previous sections except that during training the encoder of a $q$-ary ECNN outputs $q$ bits, i.e., $\bz_n\in\Real^{q},n=0,\ldots,N-1$. To decode, we may convert the $q$-ary code matrix $\bM$ into its binary version and then apply the decoding processing (same to the binary ECNN) to decode. More details about $q$-ary ECNN implementation and experiments can be found in the supplementary material.

\section*{Experiments}
\label{sect:exper}

In this section, we evaluate the robustness of ECNN under the adversarial attacks with different attack parameters. The details of these adversarial attacks are provided in the supplementary material. 
We also discuss some recently developed ensemble methods in the supplementary material, among which we experimentally compare our proposed ECNN with 1) the ECOC-based DNN proposed in \cite{VermaNIPS2019} as it shares a similar concept as ours and 2) ADP-based ensemble method proposed in \cite{PangICML2019} as it is the only method among the aforementioned ones that trains the ensemble in an end-to-end fashion. Another reason for choosing these two methods as baseline benchmarks is their reported classification accuracies under adversarial attacks are generally better than the other ensemble methods. Due to the page limitation, more experimental results can be found in the supplementary material. \footnote{Our experiments are run on a GeForce RTX 2080 Ti GPU.}

\subsection*{Setup}
\label{subsect:setup}
We test on two standard datasets: MNIST \cite{LecunPIEEE1998}, CIFAR-10 and CIFAR-100 \cite{KriTR2009}. For the encoder of ECNN, we constrain $h_n=h$ for $n=0,\ldots,N-1$ and $g_{\theta_n}=g^{\{2\}}_{n} \circ g^{\{1\}}$  for $n=0,\ldots,N-1$ so that each composite function becomes $h\circ g^{\{2\}}_{n} \circ g^{\{1\}}$. We use ResNet20 \cite{HeCVPR2016} to construct $h\circ g^{\{2\}}_{n} \circ g^{\{1\}}$. To recap, ResNet20 consists of three stacks of residual units where each stack contains three residual units, so there are nine residual units in ResNet20.  With the intent of maintaining low computational complexity, we construct the shared feature extraction function $g^{\{1\}}$ using the first eight residual units in ResNet20 while leaving the last one to $g^{\{2\}}$. The shared prediction function $h$ is a simple Dense layer. The detailed structure of ECNN is available in the supplementary material.
In the following, we use ${\rm ECNN}^{N}_{\gamma}$ to denote an ECNN, trained using a trade-off  parameter $\gamma$ used in \cref{eq:enc_dec}, with $N$ binary classifiers.  
In all our result tables, bold indicates the best performer in a particular row.



\subsection*{Performance Under White-box Attacks}
\label{subsect:whitebox}
White-box adversaries have knowledge of the classifier models, including training data, model architectures and parameters. We test the performance of ECNN in defending against white-box attacks. The default parameters used for different attack methods are provided in the supplementary material.
We compare ECNN with two state-of-the-art ensemble methods: 
\begin{enumerate}[leftmargin=\parindent,align=left,labelwidth=\parindent,labelsep=0pt]
\item The adaptive diversity promoting (ADP) ensemble model, proposed by \cite{PangICML2019}, for which we use the same architecture and model parameters as reported therein.  Specifically, we use ${\rm ADP}_{2,0.5}$ with three ResNet20s being its meta classifiers. Note that the optimal solution of ADP is attained when $M-1$ is divisible by the number of base classifiers $N$. For $M=10$, using $N=3$ is optimal for ADP and increasing $N$ beyond that does not improve its performance.
\item The ECOC-based neural network proposed by \cite{VermaNIPS2019}. We choose the most robust model named TanhEns16 reported therein, which stands for an ensemble model where the $\tanh$ function is applied element-wise to the logits, and a Hadamard matrix of order 16 is used.  As reported in \cite{VermaNIPS2019}, the TanhEns16 splits the whole network into four independent subnets and each outputs 4-bit codeword.
\end{enumerate}

When generating adversarial examples, as pointed out by \cite{Tramr2020OnAA} there are some precautions that should be taken: 1) at the decoder, we avoid taking the $\log$ of the logits before feeding them into the softmax function because taking $\log$ is numerically unstable, which leads to weak adversarial examples, and 2) we replace the softmax cross entropy loss function in PGD attack with the hinge loss proposed by  \cite{CarliniISSP2017} in order to stabilize the process of crafting adversarial examples.

The classification results on MNIST are shown in \cref{tab:adv_mnist}. We can see that while maintaining the state-of-the-art accuracy on normal images, ${\rm ECNN}^{30}_{0.1}$ improves the adversarial robustness as compared to the other two methods. For the most effective attack in this experiment, i.e., PGD attack, ECNN shows a $88.4\%-79.2\%=9.2\%$ improvement over TanhEns16 and a $88.4\%-0.2\%=88.2\%$ improvement over ${\rm ADP}_{2,0.5}$. In terms of the number of trainable parameters, ${\rm ECNN}^{30}_{0.1}$ uses $\frac{490,209-401,168}{401,168}=22.2\%$ more parameters than TanhEns16 and $\frac{818,334-490,209}{818,334}=40.1\%$ less than ${\rm ADP}_{2,0.5}$.


\begin{table}[t]
\centering
\small
\begin{tabular}{ccccccc} 
\toprule
Attack & Para.                          & ${\rm ADP}_{2,0.5}$   &   TanhEns16 & ${\rm ECNN}^{30}_{0.1}$ \\
\midrule
None  & -                               & {\bf 99.7}                  &     99.5              &    99.4        \\ 
\multirow{1}{*}{PGD}  &  $\epsilon=0.3$ & 0.2                   &        79.2           &    {\bf 88.4}      \\   
\multirow{1}{*}{C\&W} &  $\kappa=1$    &  87.1                  &        97.0          &   {\bf 99.4}    \\ 
\multirow{1}{*}{BSA} & $\alpha=0.8$      & 51.0                  &     95.0               & {\bf 99.3}   \\ 
\multirow{1}{*}{JSMA} & $\gamma=0.6$      & 1.6                  &     84.2               & {\bf 99.0}   \\ 
\# params & -              &  818,334              &  401,168          & 490,209   \\
\bottomrule
\end{tabular}
\caption{Classification accuracy (\%) on adversarial MNIST examples.} \label{tab:adv_mnist}
\end{table}
For CIFAR-10, we see from \cref{tab:adv_cifar10} that ${\rm ADP}_{2,0.5}$ has the best classification accuracy on normal examples but it fails to make any reasonable predictions under adversarial attacks. This is mainly due to the use of strong attack parameter settings. 
ECNN is consistently more robust than the competitors under different adversarial attacks. In particular, ${\rm ECNN}^{30}_{0.1}$ achieves an absolute percentage point improvement over ${\rm ADP}_{2,0.5}$ of up to $76.4\%$ (for C\&W) and over TanhEns16 of $8.5\%$ (for PGD) to $17.4\%$ (for BSA) for different attacks. Moreover, the number of trainable parameters used in ECNN is $\frac{490,497}{819,198}=59.88\%$ of that used in ${\rm ADP}_{2,0.5}$ and $\frac{490,497}{2,313,104}=21.21\%$ of that used in TanhEns16.


\begin{table}[t]
\centering
\small
\begin{tabular}{ccccccc} 
\toprule
Attack & Para.         & ${\rm ADP}_{2,0.5}$ & TanhEns16  & ${\rm ECNN}^{30}_{0.1}$   \\
\midrule
None & -               & {\bf 93.4}          & 87.5       & 85.1                     \\ 
PGD  & $\epsilon=0.04$ & 2.5                 & 63.1       & \bf{71.6}                     \\  
C\&W & $\kappa=1$      & 4.4                 & 68.0         & \bf{80.8}                         \\
BSA  & $\alpha=0.8$    & 4.3                 & 61.3         & \bf{78.7}                          \\
JSMA  & $\gamma=0.2$   & 15.8                & 68.2         & \bf{84.2}                          \\
\# params & &  819,198          & 2,313,104  & 490,497         \\
\bottomrule
\end{tabular}
\caption{Classification accuracy (\%) on adversarial CIFAR-10 examples.} \label{tab:adv_cifar10}
\end{table}

For CIFAR-100, \cref{tab:adv_cifar100} shows that the most effective attack causes the classification accuracy to drop relatively by $40.7\%=\frac{61.3-36.3}{61.3}$ for ${\rm ECNN}^{80}_{0.02}$ and by $91.5\%=\frac{62.2-5.3}{62.2}$ for ResNet20. Due to limited computational resources, we restrict ourselves to using ResNet20 as the base classifiers in ECNN. We believe that replacing ResNet20 with a more sophisticated neural network will lead to better classification accuracy on both normal and adversarial examples.
\begin{table}[t]
\centering
\small
\begin{tabular}{cccc} 
\toprule
Attacks & Para.        & ResNet20    & ${\rm ECNN}^{80}_{0.02}$ \\
\midrule
None         & -    & 62.2          & \bf{61.3}  \\ 
PGD  & $\epsilon=0.04$ & 5.3         &  \bf{36.3}     \\ 
BIM & $\epsilon=0.04$ &6.3           & \bf{42.9} \\
C\&W & $\kappa=1$      & 12.4        & \bf{52.0} \\
\bottomrule
\end{tabular}
\caption{Classification accuracy (\%) on adversarial CIFAR-100 examples.} \label{tab:adv_cifar100}
\end{table}

ECNN is compatible with many defense methods such as adversarial training (AdvT).
Experimental results with AdvT are given in \cref{tab:advT}, where AdvT augments the model's orignal loss (e.g., \cref{eq:enc_dec} for ECNN) caused by normal training examples with the loss caused by adversarial examples in each mini-batch. The ratio of adversarial examples and normal ones in each mini-batch is 1:1. In the training phase, we use PGD, where softmax cross-entropy loss is used, at $\epsilon=0.04$ with 50 iterations to craft adversarial examples. 
In the testing phase, we run PGD at $\epsilon=0.03$ for 200 iterations to attack.
As can be observed, ECNN itself is superior to pure AdvT. ECNN+AdvT achieves the best performance. 
\begin{table}[b]
\centering
\small
\begin{tabular}{cccc}
\toprule
Attacks            & Para.             & ResNet20 & ResNet20 + AdvT   \\
\midrule                     
\multirow{3}{*}{PGD} & \multirow{3}{*}{$\epsilon=0.03$} & 10.9      & 51.9               \\
\cmidrule(){3-4}
                   &              & ${\rm ECNN}^{30}_{0.1}$ & ${\rm ECNN}^{30}_{0.1}$+AdvT \\
\cmidrule(){3-4}
                 &                &59.6      & 69.4  \\
\bottomrule
\end{tabular}
\caption{Classification accuracy (\%) on adversarial CIFAR-10 examples.} \label{tab:advT}
\end{table}

\subsection*{Performance Under Black-box Physical World Attack}
\label{subsect:phy}
\begin{figure}[t]
     \centering
     \subfloat[][Normal examples]{\includegraphics[width=.40\linewidth]{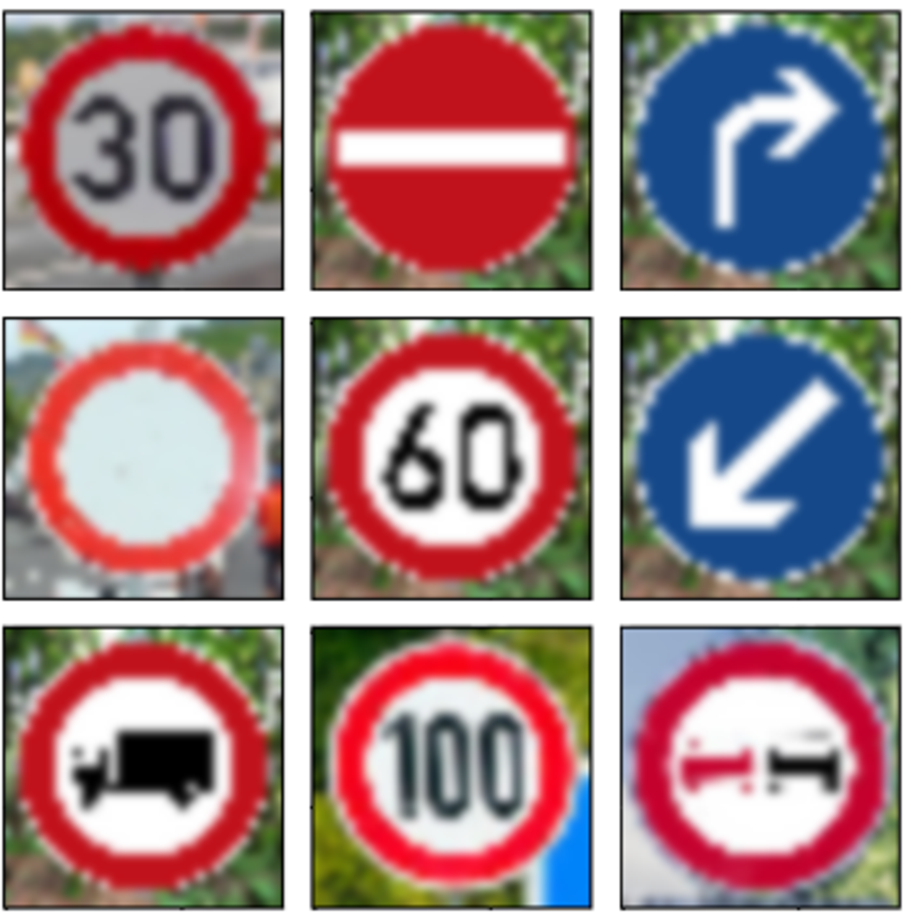}\label{fig:clean_gts}}
     \hspace{0.1in}
     \subfloat[][Adversarial examples]{\includegraphics[width=.40\linewidth]{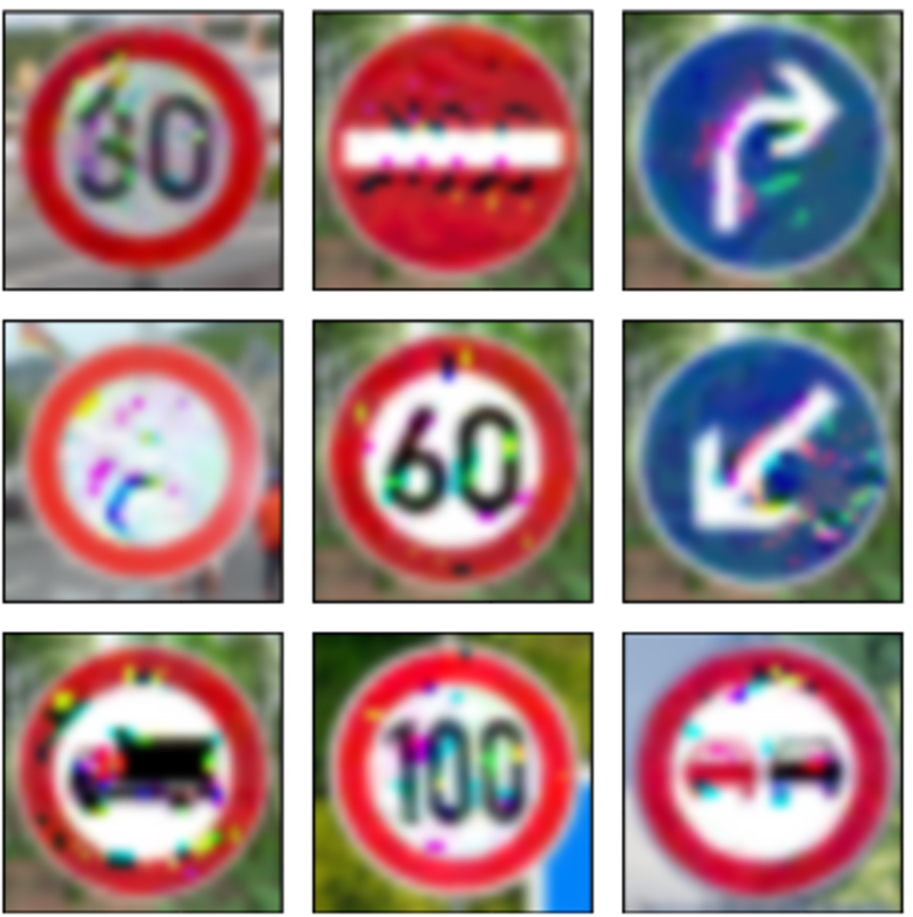}\label{fig:adv_gts}}
     \caption{German traffic examples where the adversarial examples are crafted by \cite{SitawarinDLS2018}.}
     \label{fig:gts}
\end{figure}

We conduct tests on the German Traffic Sign Recognition Benchmark (GTSRB) \cite{Stallkamp2012} under black-box setting, where adversaries do not know the model internal architectures or training parameters. An adversary crafts adversarial examples based on a substitute model and then feed these examples to the original model to perform the attack. We train ResNet20 and ECNN on the GTSRB and test them on 12630 normal examples and 392 adversarial examples  crafted by OptProjTran method 
\footnote{
The adversarial examples generated by this attack are available at \url{https://github.com/inspire-group/advml-traffic-sign}.} proposed in \cite{SitawarinDLS2018} using a custom multi-scale CNN \cite{SermanetIJCNN2011}.  The traffic sign examples are shown in \cref{fig:clean_gts} and \cref{fig:adv_gts}. 
The classification results are summarized in \cref{tab:adv_GTSRB}.
\begin{table}[t]
\centering
\small
\begin{tabular}{ccccccc} 
\toprule
Attack & ResNet20 & ${\rm ECNN}^{30}_{0.2}$   \\
\midrule
None   & 95.5     & \bf{97.2}            \\ 
OptProjTran    & 48.2              &  \bf{79.6}                     \\  
\bottomrule
\end{tabular}
\caption{Classification accuracy (\%) on 392 adversarial GTSRB examples by \cite{SitawarinDLS2018}.} \label{tab:adv_GTSRB}
\end{table}

\subsection*{Impact of Ensemble Diversity}
\label{subsect:div}
When removing the ensemble diversity from the optimization, the performance decreases significantly. Specifically, using the same shared front network for a fair comparison, we obtain $41.2\%$ accuracy after applying PGD attack using ${\rm ECNN}^{10}_{0.1}$ (cf.\ Table 2 of supplementary) on CIFAR-10, but we only achieve $17.0\%$ accuracy using ${\rm ECNN}^{10}_{0}$. 

\subsection*{Transferability Study}
\label{subsect:transfer}
\begin{figure}[t]
     \centering
     \subfloat[][w/ ens. divers., $\gamma=0.1$]{\includegraphics[width=3.3cm, height=3.2cm]{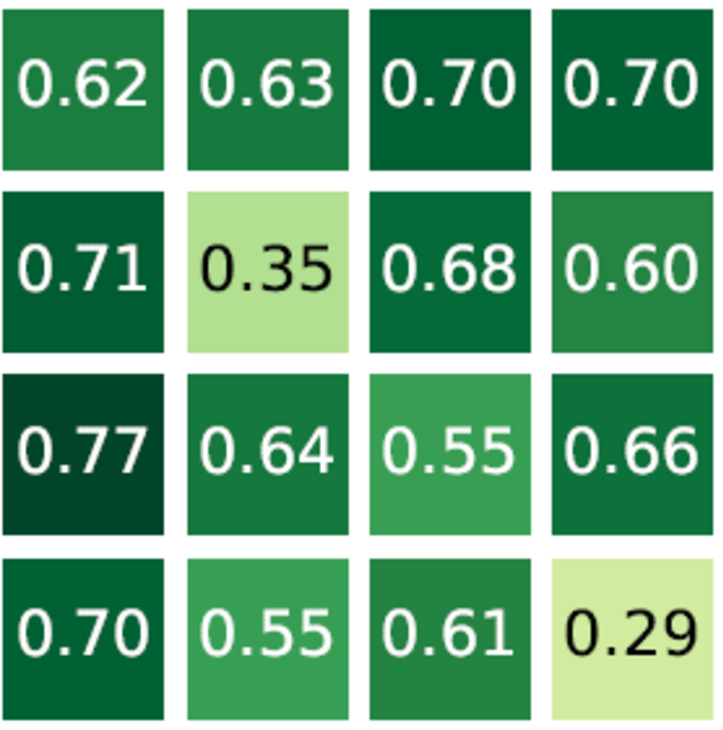}\label{fig:div}}
     \hspace{0.1in}
     \subfloat[][w/o ens. divers., $\gamma=0$]{\includegraphics[width=4.3cm, height=3.2cm]{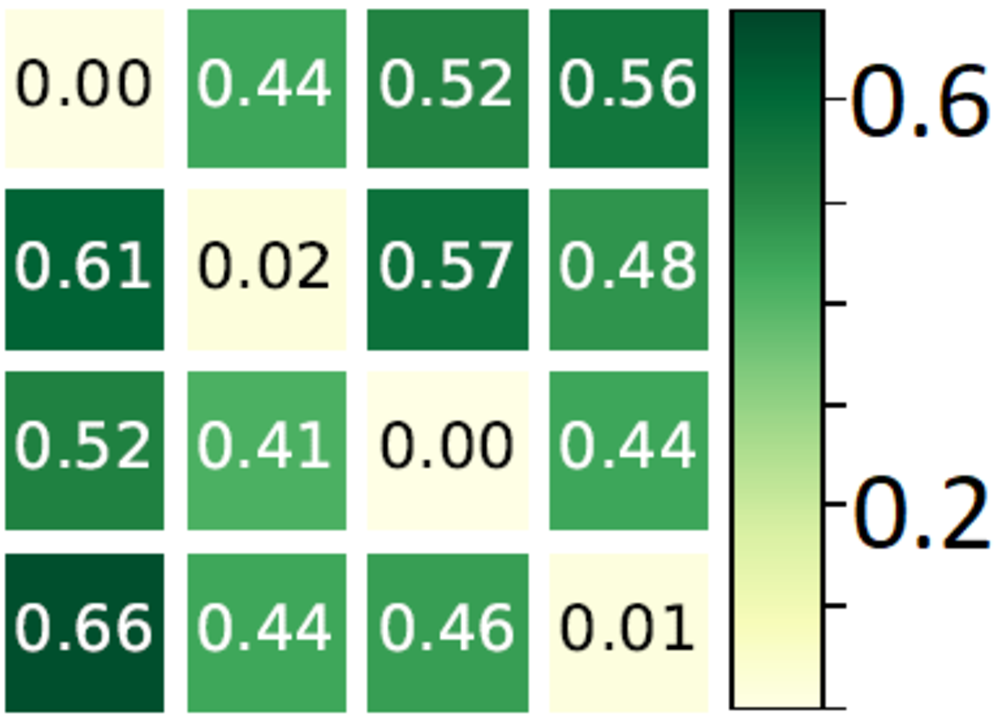}\label{fig:no_div}}
     \caption{Adversarial transferability (binary classification accuracy) among binary classifiers in ${\rm ECNN}^{10}_{\gamma}$ on CIFAR-10. The transferability among the first four networks is shown.}
     \label{fig:transfer}
\end{figure}

Transferability study is carried out using ${\rm ECNN}^{10}_{\gamma}$ on CIFAR-10, where the $(i,j)$-th entry shown in \cref{fig:transfer} is the classification accuracy using the $i$-th network as the substitute model to craft adversarial examples by running PGD at $\epsilon=0.04$ for $200$ iterations and feeding to the $j$-th network. It can be seen that 1) ECNN training yields low transferability among the binary classifiers and 2) having diversity control improves the robustness on individual networks. 

\subsection*{Impact of Parameter Sharing}
\label{subsect:para_share}
ECNN performs parameter sharing among binary classifiers at both ends of the encoder. Sharing $g^{\{1\}}$ is  inspired by the lesson from transfer learning \cite{YosNIPS2014} that features are transferable between neural networks when they are performing similar tasks. Due to \cref{lem:bilinear}, sharing $h$ is to avoid the features extracted from different binary classifiers to be permutations of each other for the same sample $x$. \cref{tab:share} shows that performing parameter sharing yields better classification accuracy on normal and adversarial examples than the one with no parameter sharing, i.e., each binary classifier is a composite function $h_n\circ g^{\{2\}}_{n} \circ g^{\{1\}}_{n}$. This is mainly because the latter which contains too many parameters overfits the data.
\begin{table}[t]
\centering
\small
\begin{tabular}{cccc} 
\toprule
Attack & Para.         & w/ para. share & w/o para. share  \\
\midrule
None & -               & 85.1      &     79.2               \\ 
PGD  & $\epsilon=0.04$ &  71.6      &     51.1                       \\  
C\&W & $\kappa=1$  &  80.8       &     73.2                       \\
\# params & - & 490,497    &   8,194,590        \\
\bottomrule
\end{tabular}
\caption{Classification accuracy (\%) on adversarial CIFAR-10 examples using ${\rm ECNN}^{30}_{0.1}$ w/ and w/o parameter sharing.} \label{tab:share}
\end{table}

\section*{Conclusion}
\label{sect:conc}
In this paper, we have presented a robust neural network ECNN that is inspired by error-correcting codes and analyzed its properties and proposed a training method that trains the binary classifiers jointly. 
Designing a code matrix by optimizing both the Hamming distance between rows and VI distance between columns makes ECNN more robust against adversarial examples. We found that performing parameter sharing at two ends of the ECNN's encoder improves ensemble's robustness while significantly reducing the number of trainable parameters. 

\section*{Acknowledgements}
This research is supported in part by A*STAR under its RIE2020 Advanced Manufacturing and Engineering (AME) Industry Alignment Fund – Pre Positioning (IAF-PP) (Grant No. A19D6a0053) and Industry Alignment Fund (LOA Award I1901E0046). The computational work for this article was partially performed on resources of the National Supercomputing Centre, Singapore (https://www.nscc.sg).

\begin{small}
\bibliography{adv_dnn}
\end{small}


\appendix

\section{Related Work and Adversarial Attacks}
\label{sect:related_app}
In this section we brief more recent works that defend against adversarial examples using ensembles of models \cite{Abb2017, Xu2018NDSS, PangICML2019,VermaNIPS2019,Sen2020empir} with more details. Then, we discuss common adversarial attacks. 

\subsection{Ensemble Models}
\label{subsect:relatedmodels}
Based on the observation that classifiers can be misled to always decide for a small number of predefined classes through adding perturbations, the work \cite{Abb2017} builds an ensemble model that consists of a generalist classifier (which classifies among all $M$ classes) and a collection of specialists (which classify among subsets of the classes). The $m$-th specialist, where $m=0,\ldots,M-1$, classifies among a subset of the original classes, denoted as $\calU_m$, within which class $m$ is most often confused in adversarial examples. The $(m+M)$-th specialist, where $m=0,\ldots,M-1$, classifies among a subset $\calU_{m+M}$ which is complementary to $\calU_m$, i.e., $\calU_{m+M}=\{0,\ldots,M-1\}\setminus \calU_m$. Altogether there are $2M$ specialists and one generalist. To classify an input, if there exists a class $m$ such that the generalist and the $M$ specialists that can classify $m$ all agree that the input belongs to class $m$, the prediction is made by averaging the outputs of the generalist and the $M$ specialists. Otherwise, at least one classifier has misclassified the input, and the prediction is made by averaging the outputs of all $2M+1$ classifiers in the ensemble.

It has also been observed that the feature input spaces are often unnecessarily large, which leaves an adversary extensive space to generate adversarial examples. The reference \cite{Xu2018NDSS} proposes to ``squeeze" out the unnecessary input features. More specifically, \cite{Xu2018NDSS} trains three independent classifiers, respectively on three different versions of an image: the original image, the reduced-color depth version and the spatially smoothed version of the original image. In the testing phase, it compares the softmax probability vectors across these three classifiers' outputs in terms of $L_1$ distance and flags the input as adversarial when the highest pairwise $L_1$ distance exceeds a threshold. This method is mainly used for adversarial detection.

Motivated by studies \cite{Pang2018a,Pang2018b} that show theoretically that promoting diversity among learned features of different classes can improve adversarial robustness, the paper \cite{PangICML2019} trains an ensemble of multiple base classifiers with an adaptive diversity promoting (ADP) regularizer that encourages diversity. The ADP regularizer forces the non-maximal prediction probability vectors, i.e., the vector containing all prediction probabilities except the maximal one, learned by different base classifiers to be mutually orthogonal, while keeping the prediction associated with the maximal probability be consistent across all base classifiers.  

Inspired by ECOC, which is mostly used with decision trees or shallow neural networks \cite{DietJAIR1994,CarcTEC2008}, the authors in \cite{VermaNIPS2019} design a DNN using error-correcting codes. It splits the binary classifiers into several groups and trains them separately. Each classifier is trained on a different task according to the columns of a predefined code matrix $\bM\in\Real^{M\times N}$ such as a Hadamard matrix with the aim to increase the diversity of the classifiers so that adversarial examples do not transfer between the classifiers.  To generate final predictions, \cite{VermaNIPS2019} proposes to map logits to the elements of a codeword and assigns probability to class $k$ as 
\begin{align*}
p_\nu(k)=\frac{\max(\boldm_k\T\nu(\bz),0)}{\sum_{i=0}^{N-1} \max(\boldm_k\T\nu(\bz),0)},
\end{align*}
where $\nu$ can be either logistic function if $\bM$ takes values in $\{0,1\}$ or $\tanh$ function if $\bM$ takes values in $\{-1,1\}$, and $\boldm_k$ denotes the $k$-th row of $\bM$.

Studies \cite{Galloway2017,Rakin2018} have suggested that quantized models demonstrate higher robustness to adversarial attacks. The work \cite{Sen2020empir} therefore independently trains multiple models with different levels of precision on their weights and activations, and makes final predictions using majority voting. For example, it uses an ensemble consisting of three models: one full-precision model, i.e., using 32-bit floating point numbers to represent activations and weights, one model trained with 2-bit activation and 4-bit weights, and another model with 2-bit activations and 2-bit weights.

Among the aforementioned ensemble methods, we experimentally compare our proposed ECNN with 1) the ECOC-based DNN proposed in \cite{VermaNIPS2019} as it shares a similar concept as ours and 2) ADP-based ensemble method proposed in \cite{PangICML2019} as it is the only method among the aforementioned ones that trains the ensemble in an end-to-end fashion. Another reason for choosing these two methods as baseline benchmarks is their reported classification accuracies under adversarial attacks are generally better than the other ensemble methods. 

\subsection{Overview of Adversarial Attacks}
\label{subsect:adv_attack}

For a normal image-label pair $(x,y)$ and a trained DNN $f_\theta$ with $\theta$ being the trainable model parameters, an adversarial attack attempts to find an adversarial example $x'$ that remains in the $L_p$-ball with radius $\epsilon$ centered at the normal example $x$, i.e., $\|x-x'\|_p\leq \epsilon$, such that $f_\theta(x') \neq y$. In what follows, we briefly present some popular adversarial attacks that will be used to verify the robustness of our proposed ECNN.

\begin{enumerate}[a)]
\item \textbf{ Fast Gradient Sign Method (FGSM)} \cite{GoodICLR2015} perturbs a normal image $x$ in its $L_\infty$ neighborhood to obtain 
\begin{align*}
x'=x+\epsilon \cdot {\rm sign}\left(\nabla_x \calL(f_\theta(x),y)\right),
\end{align*} 
where $\calL(f_\theta(x),y)$ is the cross-entropy loss of classifying $x$ as label $y$, $\epsilon$ is the perturbation magnitude, and the update direction at each pixel is determined by the sign of the gradient evaluated at this pixel.
FGSM is a simple yet fast and powerful attack. 

\item \textbf{ Basic Iterative Method (BIM)} \cite{KurakinICLR2017a} iteratively refines the FGSM by taking multiple smaller steps $\alpha$ in the direction of the gradient. The refinement at iteration $i$ takes the following form:
\begin{align*}
x'_i=x'_{i-1} + {\rm clip}_{\epsilon,x}\left(\alpha \cdot {\rm sign}\left(\nabla_x \calL(f_\theta(x),y)\right)\right),
\end{align*}
where $x'_0=x$ and ${\rm clip}_{\epsilon,x}(x')$ performs per-pixel clipping of the image $x'$. For example, if a pixel takes value between 0 and 255, 
${\rm clip}_{\epsilon,x}(x')=\min\left\{255,x+\epsilon,\max\{0,x-\epsilon,x'\}\right\}$.

\item \textbf{ Projected Gradient Descent (PGD)} \cite{MadryICLR2018} has the same generation process as BIM except that PGD starts the gradient descent from a point $x'_0$ chosen uniformly at random in the $L_\infty$-ball of radius $\epsilon$ centered at $x$.

\item \textbf{ Jacobian-based Saliency Map Attack (JSMA)} \cite{PapernotESSP} is a greedy algorithm that adjusts one pixel at a time. Given a target class label $t\ne y$, at each pixel $x_{p,q}$, where $p\in\{1,\ldots,P\}$ and $q\in\{1,\ldots,Q\}$, of a $P\times Q$ image $x$, JSMA computes a saliency map: 
\begin{align*}
S(p,q)=-\alpha_{p,q}\cdot \beta_{p,q}\cdot(\alpha_{p,q}>0)\cdot(\beta_{p,q}<0),
\end{align*}
where $\alpha_{p,q}=\frac{\partial Z(x,t)}{\partial x_{p,q}}$, $\beta_{p,q}=\sum_{j\neq t}\frac{\partial Z(x,j)}{\partial x_{p,q}}$, and $Z(x,i)$ denotes the logit associated with label $i$. The logits are the inputs to the softmax function at the output of the DNN classifier. A bigger value of $S(p,q)$ indicates pixel $(p,q)$ has a bigger impact on mis-labeling $x$ as $t$.  Let $(p^*,q^*)=\underset{p,q}{\argmax} \ S(p,q)$ and set the pixel $x_{p^*,q^*}=\min\left({\rm clip}_{\rm max}, x_{p,q}+\theta\right)$, where ${\rm clip}_{\rm max}$ is a pre-defined clipping value. This process repeats until either the misclassification happens or it reaches a pre-defined maximum number of iterations. Note that JSMA is a targeted attack by design. Its untargeted attack variant selects the target class at random.

\item \textbf{ Carlini \& Wagner (C\&W)} \cite{CarliniISSP2017}:  
In this paper, we consider the C\&W $L_2$ attack only. Let $x'=x+\frac{1}{2}(\tanh(\omega)+1)$, where the operations are performed pixel-wise. For a normal example $(x,y)$, C\&W  finds the attack perturbation by solving the following optimization problem
\begin{align*}
\underset{\omega}{\min} \ \left\| \frac{1}{2}(\tanh(\omega)+1)-x \right\|^2_2+c\cdot \ell\left(\frac{1}{2}(\tanh(\omega)+1)\right),
\end{align*}
where $\ell(x')=\max\left(Z(x',y)-\max\{Z(x',i):i\neq y\}+\kappa, 0\right)$. A large confidence parameter $\kappa$ encourages misclassification.
\end{enumerate}

\section{Proof for \cref{lem:min}}
\label{sect:proof_lem:min}

\begin{enumerate}[(a)]
\item If $\theta_n=\theta$ for all $n=0,\ldots,N-1$, we have ${\boldf}_{n,x_k}=g_{\theta_n}(x_k)=g_{\theta}(x_k)=\tilde{\boldf}_{x_k}$ for some $\tilde{\boldf}_{x_k}\in \Real^F$.  We then obtain
\begin{align}\label{eq:linear_f}
&\bphi_n \left[\tilde{\boldf}_{x_0},\ldots, \tilde{\boldf}_{x_{K-1}}\right]
= 
\left[z_{n,x_0}, \cdots, z_{n,x_{K-1}}\right], \\
&~\mbox{for all } n=0,\ldots,N-1, \nonumber
\end{align}
where $z_{n,x_k}$ denotes the logits associated with the $k$-th sample and learned by the $n$-th binary classifier.

We choose $z_{n,x_k}$ be such that the loss of \cref{eq:enc} can be made arbitrarily small. Here, we consider two loss functions:
\begin{enumerate}[(a)]
\item The cross entropy loss is used in \cref{eqn:loss}. Let $\bzeta_{n,x_k}$ be $\bzeta_{n}$ when the input sample is $x_k$. Suppose we have
\begin{align*}
\bzeta_{n,x_k}=\begin{bmatrix} 1-\max\left(\varepsilon, \xi\bM(y(x_k), n)\right) \\ \max\left(\varepsilon, \xi\bM(y(x_k), n)\right)\end{bmatrix},
\end{align*}
where $\varepsilon$ and $\xi$ are positive constants. Then, the loss of \cref{eq:enc} can be made arbitrarily small by letting $\varepsilon \to 0$ from above and $\xi\to 1$ from below. Since $\bzeta_{n,x_k}=\left[1-\sigma(z_{n,x_k}), \sigma(z_{n,x_k})\right]\T$, this is equivalent to choosing $z_{n,x_k}=\log\left(\frac{\max\left(\varepsilon, \xi\bM(y(x_k), n)\right)}{1-\max\left(\varepsilon, \xi\bM(y(x_k), n)\right)}\right)$. 
\item The hinge loss is used in \cref{eqn:loss}. The loss can be made arbitrarily small by choosing $z_{n,x_k}$ be such that $\sign\left(2\bM(y(x_k)-1, n)\right)$.
\end{enumerate}

It now suffices to show that with probability one, there exists a $\bphi_n$ solving \cref{eq:linear_f} for this choice of $\braces{z_{n,x_k} : k=0,\ldots,K-1}$.

From \cref{pri_feature}, $\braces{\tilde{\boldf}_{x_k} : k=0,\ldots,K-1}$ are almost surely linearly independent since $K\leq F$ and the joint distribution of $\{\bn_{n,x_k} : k=0,\ldots,K-1\}$ is absolutely continuous \gls{wrt} Lebesgue measure. We thus have 
\begin{align*}
\rank\begin{bmatrix} \tilde{\boldf}_{x_0} & \cdots & \tilde{\boldf}_{x_{K-1}} \\ z_{n,x_0} & \cdots & z_{n,x_{K-1}}\end{bmatrix} = 
\rank\begin{bmatrix} \tilde{\boldf}_{x_0} & \cdots & \tilde{\boldf}_{x_{K-1}}\end{bmatrix}, 
\end{align*}
i.e., each row in $\left[z_{n,x_0}, \ldots, z_{n,x_{K-1}}\right]$ belongs to the row space of $\left[\tilde{\boldf}_{x_0},\ldots, \tilde{\boldf}_{x_{K-1}}\right]$. Hence, there exists a solution $\bphi_n$ to \cref{eq:linear_f} for each $n$. The proof of \ref{it:lem2_a} is now complete.

\item If $\bphi_n=\bphi$ for all $n=0,\ldots,N-1$, we have
\begin{align}\label{eq:linear_f3}
&\bphi\left[{\boldf}_{0,x_0},\ldots, {\boldf}_{0,x_{K-1}},\ldots,{\boldf}_{N-1,x_0},\ldots, {\boldf}_{N-1,x_{K-1}}\right] \nonumber\\
&= 
\left[z_{0,x_0}, \ldots, z_{N-1,x_{K-1}}\right].
\end{align}
Since $NK\leq F$, a similar argument as the proof of \ref{it:lem2_a} can be applied here to obtain the claim.

\item If $\theta_n=\theta$ and $\bphi_n=\bphi$ for all $n=0,\ldots,N-1$, using the same notations as in the proof of \ref{it:lem2_a}, we have
\begin{align}\label{eq:linear_f5}
&\bphi \left[\tilde{\boldf}_{x_0},\ldots, \tilde{\boldf}_{x_{K-1}},\ldots,\tilde{\boldf}_{x_0},\ldots, \tilde{\boldf}_{x_{K-1}}\right] \nonumber\\
&= 
\left[z_{0,x_0}, \ldots, z_{N-1,x_{K-1}}\right],
\end{align}
which implies that $\bphi\tilde{\boldf}_{x_k}=z_{n,x_k}$ holds for all $n=0,\ldots,N-1$. This contradicts the fact that the optimal $z_{n,x_k}$, $n=0,\ldots,N-1$, are distinct because the rows in the code matrix $\bM$ are designed to be different. The proof of \ref{it:lem2_c} is now complete.
\end{enumerate}

\section{Proof for \cref{lem:bilinear}}
\label{sect:proof_lem:bilinear}

Let $z_n^y = \bphi_n (\boldf_n^y)$ denote the logits associated with class $y$ for the $n$-th binary classifier.
\begin{enumerate}[(a)]
\item
Using the same construction as in the proof of \cref{lem:min}\ref{it:lem2_a}, we choose $\{z_n^y\}$ such that the loss of \cref{eq:enc} is arbitrarily small. It suffices to show that there exists $\bphi_n$ that satisfy:
\begin{align}\label{eq:linear_f_switch}
\bphi_n\bS_n\left[{\boldf}^0,\ldots, {\boldf}^{M-1}\right]
= 
\left[z_n^0, \cdots, z_n^{M-1}\right],
\end{align}
for $n=0,\ldots,N-1$. A similar argument as that in the proof of \cref{lem:min} shows this since $\bS_n$ is invertible.  

\item
We show the following equations have no solution $\bphi$ for almost every $\{\boldf^y\}_{y=0}^{M-1}$:
\begin{align}\label{eq:bilinear_f_switch}
\bphi \bS_n{\boldf}^y= z_{n}^y,\ n=0,\ldots,N-1.
\end{align}
Similar to the proof technique used in \cite[Theorem 5.2]{johnson2014solution}, we define a map 
\begin{align*}
G: \ \Real^{FM+F} \to &  \ \Real^{NM} \nonumber \\
(\bphi\T,  {\boldf^0},\ldots, {\boldf^{M-1}}) \mapsto & \nonumber\\
&\hspace{-.6in}\left(\bphi \bS_0{\boldf^0},\ldots,\bphi \bS_0{\boldf^{M-1}},\ldots,\bphi \bS_{N-1}\boldf^{M-1}\right).
\end{align*} 
It suffices to prove that the image of $G$ has Lebesgue measure $0$ in $\Real^{NM}$. By Sard's theorem, the image of the critical set of $G$ has Lebesgue measure $0$ in $\Real^{NM}$, where  the critical set of $G$ is the set of points in $\Real^{FM+F}$ at which the Jacobian matrix of $G$, denoted as $\bJ_G\in \Real^{NM\times (FM+F)}$, has $\rank(\bJ_G)<NM$. Therefore, it suffices to show $\rank(\bJ_G) < NM$ for all points in $\Real^{FM+F}$. We have
\begin{align*}
\bJ_G = \left[\begin{array}{cccc}
(\boldf^0)\T \bS_0\T                & \bphi \bS_0        & \cdots                       & {\bf 0} \\
\vdots                               & \vdots               & \ddots                  & \vdots \\
(\boldf^{M-1})\T \bS_0\T                  & {\bf 0}              & \cdots                        & \bphi \bS_0 \\
(\boldf^0)\T \bS_1\T                 & \bphi \bS_1        & \cdots                        & {\bf 0} \\
\vdots                              & \vdots               & \ddots                  & \vdots \\
(\boldf^{M-1})\T \bS_1\T                & {\bf 0}              & \cdots                        & \bphi \bS_1 \\
\vdots                               & \vdots               & \vdots                  & \vdots \\
\end{array}
\right].
\end{align*}

It can be seen  
\begin{align*}
\bJ_G \left[\bphi, -({\boldf^0})\T, \ldots, -({\boldf^{M-1}})\T\right]\T = {\bf 0}.
\end{align*}
Therefore, the columns of $\bJ_G$ are linearly dependent and thus $\rank(\bJ_G)<FM+F\leq NM$ holds for every point in $\Real^{FM+F}$. The proof is now complete if we note that there are only finitely many permutation matrices.


\end{enumerate}

\section{Proof for \cref{lem:highPr}}
\label{sect:proof_lem:highPr}

Consider the model where we set the feature vector $\boldf_{n,x}$ to be $\boldf^{y(x)}_n$ for all $n$ and $x$. Using a similar argument as in the proof of \cref{lem:min}\ref{it:lem2_b}, it can be shown that there exists $\braces{\bphi_n=\bphi}_{n=0}^{N-1}$ such that the loss of \cref{eq:enc} is sufficiently small so that $z^{y(x_k)}_{n}(\bM(y(x_k),n))> z^{y(x_k)}_{n}(1-\bM(y(x_k),n))$, for all $k=0,\ldots,K-1$, where $z^{y(x_k)}_{n} = \bphi(\boldf^{y(x_k)}_{n})$. For this choice of $\bphi_n=\bphi$ now applied to the model in the lemma, we have for each $n$ and $x_k$,
\begin{align*}
z_{n,x_k} 
&= \bphi \boldf_{n,x_k}\\
&= \bphi (\boldf^{y(x_k)}_n + \bn_{n,x_k})\\
&= z^{y(x_k)}_{n} + \bphi \bn_{n,x_k},
\end{align*}      
so that
\begin{align*}
\sign\left(z_{n,x_k}(0)-z_{n,x_k}(1)\right)=\sign\left(z^{y(x_k)}_{n}(0)-z^{y(x_k)}_{n}(1)\right)
\end{align*}
if $\left|\bn_{n,x_k}\T\left(\bphi_{0}-\bphi_{1}\right)\right| < \left|z^{y(x_k)}_{n}(0)-z^{y(x_k)}_{n}(1)\right|$, where $\bphi=\left[\bphi_{0}\T,\bphi_{1}\T\right]\T$. From the Cauchy-Schwarz inequality, $\left|\bn_{n,x_k}\T\left(\bphi_{0}-\bphi_{1}\right)\right| \leq \|\bn_{n,x_k}\|_2 \|\bphi_{0}-\bphi_{1}\|_2$ and thus zero classification error for the $n$-th binary classifier is achievable if $\|\bn_{n,x_k}\|_2$ is sufficiently small. The proof is now complete.

\section{Proof for \cref{lem:solution}}
\label{sect:proof_lem:solution}

Suppose cross entropy loss is used in \cref{eqn:loss}. The optimization in \cref{eq:enc_dec} w.r.t. $\bzeta_{n,x_k}$ becomes 
\begin{align}\label{eq:enc_div2}
\min_{\bzeta_{n,x_k}}&~ \frac{1}{NK}\sum_{k=0}^{K-1} \sum_{n=0}^{N-1} -\left({\bf 1}_{y_n(x_k)}-\gamma \bzeta_{n,x_k}\right)\T\log\left(\bzeta_{n,x_k}\right) \\
{\rm s.t.}&~\sum_{i=0}^1 \bzeta_{n,x_k}(i)=1, \nonumber 
\end{align}
where $\gamma>0$.

The Lagrangian of \cref{eq:enc_div2} is 
\begin{align*}
L=&\frac{1}{NK}\sum_{k=0}^{K-1}\sum_{n=0}^{N-1}\left( -{\bf 1}_{y_n(x_k)}\T\log\left(\bzeta_{n,x_k}\right)+
\gamma \bzeta_{n,x_k}\T\log(\bzeta_{n,x_k}) \right) \nonumber\\
&-\sum_{k=0}^{K-1}\sum_{n=0}^{N-1} a_{n,k} \left(1-\sum_i \bzeta_{n,x_k}(i)\right),
\end{align*}
where $a_{n,k}\geq 0,\text{ for all } n,k$. Differentiating $L$ w.r.t. $\bzeta_{n,x_k}(i)$ and equating to zero yields
\begin{align*}
\frac{\partial L}{\partial \bzeta_{n,x_k}(y_n(x_k))} = &
-\frac{1}{\bzeta_{n,x_k}(y_n(x_k))}+\gamma \left(\log(\bzeta_{n,x_k}(y_n(x_k)))+1\right)\\
&+a_{n,k}
=0,\\
\frac{\partial L}{\partial \bzeta_{n,x_k}(i)} = &
\gamma \left(\log(1-\bzeta_{n,x_k}(y_n(x_k)))+1\right)+a_{n,k}
=0,\\
&\text{ for all } i\neq y_n(x_k).
\end{align*}
Equating the above two equations, we have $\gamma\log\frac{\bzeta_{n,x_k}(y_n(x_k))}{1-\bzeta_{n,x_k}(y_n(x_k))}=\frac{1}{\bzeta_{n,x_k}(y_n(x_k))}$. This completes the proof.

\section{Proof for \cref{lem:I}}
\label{sect:proof_lem:I}

We can easily see that $s(k,\ell)=\frac{1}{M}$ if $\ell\in C^k$; $s(k,\ell)=0$ otherwise. Then, $I(\by,\bc)$ in \cref{eq:I} can be rewritten as 
\begin{align*}
I(\by,\bc)&=\sum_{k=0}^{q-1} \frac{\left|C^{k}\right|}{M} \log \frac{\frac{1}{M}}{\frac{\left|C^{k}\right|}{M}\frac{1}{M}} \\
&= \sum_{k=0}^{q-1} \frac{\left|C^{k}\right|}{M} \log \frac{M}{\left|C^{k}\right|}.
\end{align*}
The maximum of $I(\by,\bc)$ is achieved when $\left|C^{k}\right|=M/q,\text{ for all } k$,  i.e., $\max I(\by,\bc)=\log q$, which increases with $q$. The proof is complete.


\section{Training $q$-ary ECNN}
\label{sect:qary_train}
We introduce a method to implement a general $q$-ary ECNN, where $q\geq 2$.
\subsection{Encoder} 

The logits, i.e., the encoder's outputs, in \cref{fig:ecn} becomes $\bz_n=h_n({\boldf}_n)\in\Real^q$. Using a multi-class hinge loss \cite{CarliniISSP2017}, the encoder can be formulated as an optimization problem:
\begin{align}\label{eq:enc_q}
\min_{\{\theta_n,\bphi_n\}_{n=0}^{N-1}}&\ \frac{1}{NK} \sum_{k=0}^{K-1} \sum_{n=0}^{N-1} 
\max\left(\max\{\bz_{n,x_k}(i):  i\neq \bM(y(x_k),n)\} \right. \nonumber \\
&\left. -\bz_{n,x_k}(\bM(y(x_k),n))+\kappa, 0\right) \\
{\rm s.t.}&~ \text{for $k=0,\ldots,K-1$ and $n=0,\ldots,N-1$}, \nn
&\ z_{n,x_k}= \bphi_n  g_{\theta_n}(x_k), \nonumber
\end{align}
where $\bM$ denotes a $q$-ary code matrix and $\kappa$ is the confidence level which is configurable.
Finally, the logits $\bz_n$ are concatenated into $\bz=\left[\bz_0\T,\ldots,\bz_{N-1}\T\right]\T\in\Real^{qN}$, which is the input to the decoder in ECNN.

\subsection{Decoder} 
The decoder maps the logits $\bz$ to the prediction probabilities $\bp={\sigma}\left(\bM \nu(\bz)\right)\in\Real^M$, where $\sigma(\cdot)$ denotes the softmax function and $\nu(\cdot)$ denotes the logistic function. 

\subsection{Joint Optimization}
The joint optimization for a $q$-ary ECNN is as follows
\begin{align}\label{eq:enc_dec_q}
\min_{\{\theta_n\}_{n=0}^{N-1},\bphi}&~ 
\frac{1}{NK}\sum_{k=0}^{K-1} \sum_{n=0}^{N-1} 
\max\left(\max\{\bz_{n,x_k}(i):i\neq \bM(y(x_k),n)\} \right. \nn
&\ \left. -\bz_{n,x_k}(\bM(y(x_k),n))+\kappa, 0\right)   - \gamma\bzeta_{n,x_k}\log\left(\bzeta_{n,x_k}\right), \\
{\rm s.t.} &~ \text{for } n=0,\ldots,N-1,\ k=0,\ldots,K-1,\nn
&~ \bz_{n,x_k}=\bphi g_{\theta_n}(x_k),\  \bzeta_{n,x_k} = \sigma(\bz_{n,x_k}), \nonumber
\end{align}
where $\gamma\geq 0$ is a configurable weight.

\section{Further Experiments}
\label{sect:fur_exp}
Denote $\text{Res}_{i,j}, i,j\in\{0,1,2\}$ as the $j$-th residual unit in the $i$-th stack. \cref{tab:func} shows how the functions $g^{\{1\}}$ and $g^{\{2\}}_{n}$ are constructed using the residual units in ResNet20. 
\begin{table}[!tbh]
\centering
\caption{The neural networks for $h, g^{\{2\}}_{n}, g^{\{1\}}$.} \label{tab:func}
\begin{tabular}{lll} 
\toprule
$g^{\{1\}}$         & $g^{\{2\}}_{n}$    &   $h$                 \\
\midrule
$\text{Res}_{0,0}$   & $\text{Res}_{2,2}$  &   Dense(1, Linear)       \\
$\cdots$            & Dense(32, ReLu)    &          \\
$\text{Res}_{2,1}$   &      &          \\
\bottomrule  
\end{tabular}
\end{table}
The ECNNs used to get all our experimental results do not use batch norms (BNs), however, we found that adding BNs to ECNN will slightly improve ECNN's classification accuracy on both normal and adversarial examples.

The PGD attack implemented in \cref{tab:adv_mnist} and \cref{tab:adv_cifar10} uses a hinge loss proposed in \cite{CarliniISSP2017}. The optimal perturbation  $\delta^*$  is thus found by solving
\begin{align*}
\min_\delta \ -
\max\left(\bs_{x+\delta}(y(x+\delta))-\max\{\bs_{x+\delta}(i):i\neq y(x)\}+c, 0\right),
\end{align*}
where $\bs$ denotes the logits from the decoder and $c=50$.

\subsection{Number of Ensemble Members in ECNN}
For a $10$-label classification problem, representing labels with unique codewords requires $N\geq 4$. In this experiment, we independently generate five code matrices for $N=10,15,20,25,30$, respectively, and use them to construct five ECNNs. The settings of adversarial attacks are the same as  \cref{tab:adv_cifar10}.  \cref{tab:ecn_acc_N} shows that when $N$ is small, ECNN has inferior performances on adversarial examples. One possible reason is that we use a dense layer with the same feature dimension $F=32$ for all $N$ in our experiments. From \cref{lem:bilinear}, we know that when $F\gg N$, it may happen that the features $\boldf^{y(x)}_n$ for different binary classifiers $n=0,\ldots,N-1$ are permutations of each other for the same sample $x$. Adversarial attacks on a particular binary classifier can then translate to the other classifiers.


\begin{table*}[!thb]
\centering
\caption{Classification accuracy (\%) on adversarial CIFAR10 examples using ${\rm ECNN}^{N}_{0.1}$ where $N=10,15,20,25,30$.} \label{tab:ecn_acc_N}
\begin{tabular}{ccccccc} 
\toprule
Attack & Para.         & $N=10$  & $N=15$   & $N=20$ & $N=25$  & $N=30$  \\
\midrule
None & -                 &  84.6   &   85.5    & 85.1   & 85.5 &   85.1           \\ 
\midrule                       
PGD  & $\epsilon=0.04$     &   41.2      &  49.9      &  50.2    &  48.5  &  71.6           \\  
\midrule    
C\&W & -                  &   72.0      &   71.4     &  74.1    & 71.8  &  80.8              \\
\midrule
\# of params & -  &  294,977   &  343,857   & 392,737   &441,617  & 490,497 \\
\bottomrule
\end{tabular}
\end{table*}

\subsection{Parameter Sharing}
As shown in \cref{tab:func}, ECNN performs parameter sharing among binary classifiers at both ends of the encoder. Sharing $g^{\{1\}}$ is  inspired by the idea from transfer learning \cite{YosNIPS2014} which says the features are transferable between neural networks when they are performing similar tasks. Due to \cref{lem:bilinear}, sharing $h$ is to avoid the features extracted from different binary classifiers to be permutations of each other for the same sample $x$. 

Define $|g^{\{1\}}|$ as the number of trainable parameters in $g^{\{1\}}$ and define $|g^{\{2\}}_{n}|,n=0,\ldots,N-1$ and $|h|$ analogously. The total number of trainable parameters in ECNN can be computed as $|g^{\{1\}}|+\sum_{n=0}^{N-1}|g^{\{2\}}_{n}|+|h|$. If $|g^{\{1\}}|+|h|$ is much larger than any $|g^{\{2\}}_{n}|$, the growth of the total number of parameters in ECNN when $N$ increases will be very slow. \cref{tab:share} shows that performing parameter sharing yields better classification accuracy on normal and adversarial examples than the one with no parameter sharing, i.e., each binary classifier is a composite function $h_n\circ g^{\{2\}}_{n} \circ g^{\{1\}}_{n}$. This is mainly because the latter which contains too many parameters overfits the data.
\begin{table}[!thb]
\centering
\caption{Classification accuracy (\%) on adversarial CIFAR10 examples using ${\rm ECNN}^{30}_{0.1}$ with and without parameter sharing.} \label{tab:share}
\begin{tabular}{cccc} 
\toprule
Attack & Para.         & w/ para. share & w/o para. share  \\
\midrule
None & -               & 85.1      &     79.2               \\ 
\midrule                       
PGD  & $\epsilon=0.04$ &  71.6      &     51.1                       \\  
\midrule    
C\&W & -               &  80.8       &     73.2                       \\
\midrule
\# of params & - & 490,497    &   8,194,590        \\
\bottomrule
\end{tabular}
\end{table}

\subsection{Code Matrix}
In this experiment, we investigate the impact of code matrix design on ECNN's performance. To highlight the benefit of optimizing both row and column distances, we construct a code matrix $\bM'=\left[\bM_a,\mathds{1}-\bM_a\right]$ of size $10\times 30$ where $\bM_a$ is of size $10\times 15$ and is the first $15$ columns extracted from the code matrix generated by optimizing \cref{eq:Mobj_alternative}. $\bM'$'s minimum VI distance between columns is 0 while its minimum Hamming distance between rows remains large. As shown in \cref{tab:codematrixdouble}, the ECNN trained on $\bM'$ has worse testing accuracy on adversarial examples so that optimizing both row and column distance is advised in practice.
\begin{table}[!thb]
\centering
\caption{Classification accuracy (\%) on adversarial CIFAR10 examples with a codematrix whose minimum VI distance between columns is 0.} \label{tab:codematrixdouble}
\begin{tabular}{cccc} 
\toprule
Attack & Para.         & $\bM$ & $\bM'$ \\
\midrule
None & -               & 85.1      &     84.0              \\ 
\midrule                       
PGD  & $\epsilon=0.04$ &  71.6      &     31.1                      \\  
\midrule    
C\&W & -               &  80.8       &     56.0                    \\
\bottomrule
\end{tabular}
\end{table}

\subsection{Fine-tuning C\&W Attack and Logits Level Attack Using PGD}
Reference \cite{zhang2020adversarial} reports that ECOC-based DNN, i.e., TanhEns16, proposed in \cite{VermaNIPS2019} is vulnerable to the white-box attacks when they are carried out properly. So, we test our ECNN's robustness under the fine-tuning C\&W attack suggested in \cite{zhang2020adversarial} and PGD attack which works on the logits level from the encoder, i.e., $\bz_n, n=0,\ldots,N-1$. For PGD attack, the hinge loss \cite{CarliniISSP2017} is used as the attack loss function. 
We use $q$-ary ECNN as the training model. \cref{tab:adv_q} shows that both attacks are not very successful to fool $q$-ary ECNN with $q=2,3,4$. One reason could be that we use the multi-class hinge loss proposed in \cite{CarliniISSP2017} as the loss function during training which counteracts these two attacks as they are  also using this hinge loss as their attack loss functions. Moreover, we suspect that the attack parameters and/or loss functions for these two attacks are not optimized. So, as a future work, we will test ECNN using more effective attacks.

\begin{table*}[!thb]
\centering
\caption{Classification accuracy (\%) on adversarial CIFAR10 examples crafted by the fine-tuning C\&W attack suggested in \cite{zhang2020adversarial} and PGD attack working on logits $\bz_n, n=0,\ldots,N-1$ at the encoder's outputs  using $q$-ary ${\rm ECNN}^{30}_{0.1}$ where $q=2,3,4$. We set the confidence level $\kappa=1$ in \cref{eq:enc_dec_q}.} \label{tab:adv_q}
\begin{tabular}{ccccccc} 
\toprule
Attack & Para.         & $q=2$ & $q=3$  & $q=4$   \\
\midrule
None & -               &  85.0  & 84.4      &  84.3               \\ 
\midrule                        
\multirow{3}{*}{Fine-tuning C\&W \cite{zhang2020adversarial}}& initial constant$=0.1$ &      &    &             \\
& max iteration $=2000$ & 81.5 &  82.5  & 82.0\\
& confidence $\kappa=0$ & &  &\\
\midrule
PGD on logits $\bz_n$ & $\epsilon=0.04$    &  79.1  & 80.2      &  80.4               \\ 
\midrule
\# of params & - &   490,530    & 490,563   &    490,596       \\
\bottomrule
\end{tabular}
\end{table*}

\end{document}